\pdfoutput=1

\documentclass[11pt]{article}

\usepackage[]{emnlp2021}

\usepackage{times}
\usepackage{latexsym}
\usepackage{booktabs,multirow,array}
\usepackage{adjustbox}
\usepackage{bm}
\usepackage{graphicx}
\usepackage{enumitem}
\usepackage{amsfonts,amsmath}

\usepackage[T1]{fontenc}

\usepackage[utf8]{inputenc}

\usepackage{microtype}

\title{Learning to Rank Question Answer Pairs with Bilateral Contrastive Data Augmentation\thanks{\hspace{2mm} The work described in this paper is substantially supported by a grant from the Asian Institute of Supply Chains and Logistics, the Chinese University of Hong Kong.}}

\author{Yang Deng, Wenxuan Zhang, Wai Lam \\
  The Chinese University of Hong Kong \\
  \texttt{\{ydeng,wxzhang,wlam\}@se.cuhk.edu.hk} \\}

\begin{document}
\maketitle
\begin{abstract}
In this work, we propose a novel and easy-to-apply data augmentation strategy, namely \textbf{Bi}lateral \textbf{G}eneration (\textbf{BiG}), with a contrastive training objective for improving the performance of ranking question answer pairs with existing labeled data. 
In specific, we synthesize pseudo-positive QA pairs in contrast to the original negative QA pairs with two pre-trained generation models, one for question generation, the other for answer generation, which are fine-tuned on the limited positive QA pairs from the original dataset. 
With the augmented dataset, we design a contrastive training objective for learning to rank question answer pairs. 
Experimental results on three benchmark datasets show that our method significantly improves the performance of ranking models by making full use of existing labeled data and can be easily applied to  different ranking models. 
\end{abstract}

\section{Introduction}

Ranking question answer pairs, also known as answer selection, is a fundamental task in question answering (QA) systems. 
It aims to rank a set of candidate answers for selecting the relevant or correct answers to the given question. 
Many efforts have been made on developing various neural models to measure the relevance degree between the question and answer pair, including Siamese Structure~\cite{hyperqa,sigir20-as}, Attention-based Structure~\cite{sigir17-poa,sigir18-kablstm,tois21}, and Compare-Aggregate Structure~\cite{cikm19-ca}. 
Recently, models with contextualized representations, e.g., ELMo~\cite{elmo} and BERT~\cite{bert}, contribute to major improvement on answer selection~\cite{cikm19-ca, tanda}.

Acquiring the ground-truth positive answer to the given question or manually annotating the relevance degree between a QA pair can be extremely time-consuming. Therefore, the existing answer selection datasets are often limited in scale. Such data scarceness issue poses a great challenge for the modern neural models, which are often data-hungry for the training process. Moreover, the negative samples, i.e., the irrelevant answers to the given question are constructed by random negative sampling from the whole document pool~\cite{yahooqa}, leaving less to learn for differentiating the relevant answer from a set of highly irrelevant answers~\cite{sigir17-hard-negative,emnlp19-hard-negative}. 
Therefore, the data scarceness and label imbalance issues lead to the underutilization of the in-domain knowledge in the original datasets, and even a waste of valuable labeled data.

\begin{table}
\centering
\setlength{\abovecaptionskip}{2pt}   
\setlength{\belowcaptionskip}{2pt}
\fontsize{9}{11}\selectfont
\begin{adjustbox}{max width=\linewidth}
\begin{tabular}{p{1.15\linewidth}}
\toprule 
$\mathbf{q}$: How many music awards has Katy Perry won?\\
\midrule
$\mathbf{a}^-_1$: Katy Perry is an American singer from Santa Barbara California.\\
$\mathbf{a}^-_2$: Perry has released three studio albums including her debut Christian effort Katy Hudson One of the Boys and Teenage Dream. \\
$\mathbf{a}^-_3$: Perry's mainstream career was kicked off with the massive chart success of her debut single I Kissed a Girl.\\
\midrule
\midrule
$\mathbf{a}^*$: Katy Perry has won 11 awards for her music career including three Grammy awards and two American Music Awards. \\
\midrule
$\mathbf{q}^*_1$: What city is Katy Perry from? \\
$\mathbf{q}^*_2$: How many albums has Katy Perry released in her career? \\
$\mathbf{q}^*_3$: Who is the singer Katy Perry?\\
\bottomrule
\end{tabular}
\end{adjustbox}
\caption{An Example from WikiQA with Synthesized Data. $\mathbf{q}, \mathbf{a}^-_1, \mathbf{a}^-_2, \mathbf{a}^-_3$ are the original data, but all these answers are irrelevant to the given question. $\mathbf{a}^*, \mathbf{q}^*_1, \mathbf{q}^*_2, \mathbf{q}^*_3$ are the corresponding synthesized data.}
\label{example}
\vspace{-0.6cm}
\end{table}

Some early works~\cite{acl17-tl,coling18-kan} leverage transfer learning (TL) based approaches to transfer knowledge from large-scale QA datasets, e.g., SQuAD~\cite{squad}, for improving resource-poor answer selection tasks. With similar motivations, \citet{cikm19-ca} combine ELMo with transfer learning from a large-scale relevant-sentence-selection dataset to enhance the performance of the compare-aggregate model.
\citet{tanda} adopt BERT-based models with a Transfer-and-Adapt (TANDA) approach by using a large and high-quality QA dataset. However, such TL-based methods heavily rely on the existence of  large-scale high-quality datasets with similar problem settings, which is also unrealistic in real-world applications. Besides, compared with utilizing cross-domain knowledge via TL, it would be more efficient to take advantage of the in-domain knowledge from the original dataset. 

Another way to handle the data scarceness and label imbalance issues is to employ hard negative sampling strategies~\cite{sigir17-hard-negative,emnlp19-hard-negative} or curriculum training approaches~\cite{sigir20-curriculum} for taking advantages of those randomly sampled negative answers. 
The essential motivation behind these methods is that the hard negative samples or the ``difficult" samples play a more important role in training a better ranking model.
However, those ``less" important samples are likely to be overlooked in such cases, leading to a potential waste of data. 
In order to maximize the utility of the existing labeled data, we aim at investigating data augmentation strategy to construct weak supervision signals for enhancing answer selection models.

In this work, we propose an effective data augmentation strategy, namely \textbf{Bi}lateral \textbf{G}eneration (\textbf{BiG}), with a contrastive training objective for improving the performance of ranking question answer pairs with existing labeled data. 
In specific, we first fine-tune a Question Generator (QG) and an Answer Generator (AG) from the pre-trained generative language model, such as BART~\cite{bart}, with the positive samples in the training data. 
Then, as the example shown in Table~\ref{example},  we use the fine-tuned QG and AG to generate a high-quality question-related answer $\mathbf{a}^*$ and answer-focused questions $\mathbf{q}^*$ from the given negative samples $(\mathbf{q},\mathbf{a}^-)$. 
The synthesized QA pairs $(\mathbf{q},\mathbf{a}^*)$ and $(\mathbf{q}^*,\mathbf{a}^-)$ can be regarded as new ``positive" samples in contrast to the original negative sample. 
By doing so, given a small amount of labeled data even with extremely imbalanced labels, such data augmentation strategy can easily synthesize high-quality data with weak supervision signals. 
Although such pseudo-positive QA pairs may not be perfectly matched, e.g., $(\mathbf{q}_3^*,\mathbf{a}_3^-)$, they are supposed to be more interrelated than the original negative QA pairs, which can be served as pairwise samples for contrastive estimation.  
For instance, $(\mathbf{q}_3^*,\mathbf{a}_3^-)$ can be used to establish a contrastive sample with $(\mathbf{q},\mathbf{a}_3^-)$, rather than be simply regarded as a positive QA pair. 
Therefore, we further design a contrastive training objective for ranking QA pairs.  

We conduct experiments by applying the proposed BiG method with both the attention-based ranking model and the pre-trained BERT ranking model. Experimental results on three benchmark datasets including TREC-QA, WikiQA, and ANTIQUE show the effectiveness and the applicability of our method, which significantly improves the original ranking models across all datasets.

\section{Method}\label{sec:method}

\subsection{Preliminaries}\label{sec:notation}
Let an ad-hoc neural ranking model be denoted as $R_\theta(\mathbf{q},\mathbf{a})$, which measures the relevance degree of a given question answer pair $(\mathbf{q},\mathbf{a})$ parameterized by the model parameters $\theta$. 
Given a set of training samples, there are three mainstream approaches to optimize the ranking performance, namely, pointwise training, pairwise training, and listwise training \cite{coling18-survey,sigir20-ar}. Here we focus on pointwise and pairwise training, since listwise training is less practical in applications when there is a large set of candidate answers. 

The pointwise training aims to optimize the predicted probability distribution of each QA pair to be consistent with the ground-truth label by minimizing the cross-entropy loss function:

\vspace{-0.3cm}
\begin{small}\begin{align}
 p_i&=\mathrm{softmax}(R_\theta(\mathbf{q}_i,\mathbf{a}_i)),\\
 \mathcal{L}^\mathrm{pt}(\mathbf{q}_i,\mathbf{a}_i,y_i)&=-\left[y_i\log{p_i}+\left(1-y_i\right)\log{\left(1-p_i\right)}\right],
\end{align}\end{small}
where $y_i$ is the label of the $i$-th training sample.

On the other hand, the pairwise training is to optimize the relative order of a pair of predicted relevance scores, i.e., $R_\theta(\mathbf{q},\mathbf{a}^+)$ and $R_\theta(\mathbf{q},\mathbf{a}^-)$, which typically adopt the same question $\mathbf{q}$ with one relevant answer $\mathbf{a}^+$ and one irrelevant answer $\mathbf{a}^-$. The main idea behind pairwise training is contrastive estimation. One widely-used pairwise loss function is the hinge loss:

\vspace{-0.3cm}
\begin{small}\begin{equation}
    \mathcal{L}^\mathrm{pr}(\mathbf{q},\mathbf{a}^+,\mathbf{a}^-) = \max(0, M - R_\theta(\mathbf{q},\mathbf{a}^+) + R_\theta(\mathbf{q},\mathbf{a}^-)),
\end{equation}\end{small}
where the model learns to rank the relevant answer $\mathbf{a}^+$ higher than the irrelevant answer $\mathbf{a}^-$, by imposing a margin $M$ between their relevance scores with the given question $\mathbf{q}$. 

\subsection{BiG Data Augmentation}
To maximize the utility of the valuable labeled data as well as alleviate the label imbalance issue, pre-trained language models, such as BART~\cite{bart}, preserve promising advantages of synthesizing high-quality pseudo QA pairs through conditional generation, which can serve as weak supervision signals for contrastive learning with original data. 
Due the nature of question answering, there are two intuitive ways to generate augmentations, \textit{i.e.}, question generation and answer generation, namely \textbf{Bi}lateral \textbf{G}eneration (\textbf{BiG}).

With the pre-trained conditional generation model, a certain number of training data is required to fine-tune a Question Generator (QG) and an Answer Generator (AG). We adopt the positive samples, i.e., the relevant $(\mathbf{q},\mathbf{a}^+)$ pair, as the training data to fine-tune the QG and AG:

\vspace{-0.3cm}
\begin{small}\begin{align}
    \mathbf{H} = \mathrm{QGEnc}(\mathbf{a}^+;\mathrm{[SEP]}),\quad
    \mathbf{q} = \mathrm{QGDec}(\mathbf{H}),\\
    \mathbf{H} = \mathrm{AGEnc}(\mathbf{q};\mathrm{[SEP]}),\quad
    \mathbf{a}^+ = \mathrm{AGDec}(\mathbf{H}),
\end{align}\end{small}
where [SEP] is a separation token.

Then, the learned QG and AG models can be employed to generate answer-related questions $\mathbf{q}^*$ and question-related answers $\mathbf{a}^*$, respectively. 
In our case, we are able to synthesize pseudo-positive QA pairs from the original negative QA pairs. For instance, given a negative QA pair. $(\mathbf{q},\mathbf{a}^-)$, two pseudo-positive QA pairs, i.e., $(\mathbf{q^*},\mathbf{a}^-)$ and $(\mathbf{q},\mathbf{a}^*)$, can be generated by the learned question generator QG and answer generator AG:

\vspace{-0.3cm}
\begin{small}\begin{gather}
\mathbf{q}^* = \mathbf{QG}(\mathbf{a}^-;\mathrm{[SEP]}),\quad
\mathbf{a}^*  = \mathbf{AG}(\mathbf{q};\mathrm{[SEP]}).
\end{gather}\end{small}
One one hand, since the generation model is pre-trained on large scale corpus and further fine-tuned on positive QA pairs, we are guaranteed to generate reliable questions or answers related to the source text. In other words, the relevance degree between the synthesized QA pairs is supposed to be sufficiently higher than the original negative QA pairs constructed by randomly sampling. 
One the other hand, compared with the original positive QA pairs, the synthesized QA pairs are likely to contain certain redundant information or lack of necessary information.
Regarding the synthesized QA pairs as ground-truth positive pairs may introduce undesired noises into the original dataset. 
Therefore, the pseudo-positive nature of the systhesized QA pairs leaves room for contrastive learning.

\subsection{Contrastive Training Objective}
As mentioned in Section~\ref{sec:notation}, conventional supervised ranking problem can only optimize the ranking performance with existing labeled data. Here we introduce contrastive learning to rank by leveraging generation-augmented contrastive data for further improving the ranking performance.

Given a question $\mathbf{q}$ from the original dataset, there are a set of relevant answers $\mathbf{A}^+ = \{\mathbf{a}^+\}$ and a set of irrelevant answers $\mathbf{A}^- =\{\mathbf{a}^-\}$, correspondingly.
The pairwise training for conventional supervised learning to rank can be defined as:

\vspace{-0.3cm}
\begin{small}
\begin{equation}\label{pair}
    \mathcal{L}_\text{set}^\mathrm{pr}(\mathbf{q},\mathbf{A}^+,\mathbf{A}^-) = \frac{\sum\limits_{\mathbf{a}^+\in \mathbf{A}^+} \sum\limits_{\mathbf{a}^-\in \mathbf{A}^-} \mathcal{L}^\mathrm{pr}(\mathbf{q},\mathbf{a}^+,\mathbf{a}^-)}{|\mathbf{A}^+|\cdot|\mathbf{A}^-|}.
\end{equation}\end{small}
\vspace{-0.3cm}

With the synthesized pseudo-positive samples, we leverage such contrastive weak supervision signals to enhance the pairwise training in Eq.\ref{pair}:

\vspace{-0.3cm}
\begin{small}\begin{equation}\label{contrastive}
\begin{split}
    &\mathcal{L}_\text{set}^\mathrm{ctrst}(\mathbf{q},\mathbf{A}^+,\mathbf{A}^-) = \mathcal{L}_\text{set}^\mathrm{pr}(\mathbf{q},\mathbf{A}^+,\mathbf{A}^-) \\
    &+ \frac{1}{|\mathbf{A}^-|} \sum_{\mathbf{a}^-\in \mathbf{A}^-} \left[\mathcal{L}^\mathrm{pr}(\mathbf{q},\mathbf{a}^*,\mathbf{a}^-) + \mathcal{L}^\mathrm{pr}(\mathbf{q}^*, \mathbf{q}, \mathbf{a}^-)\right],
\end{split}
\end{equation}\end{small}
where the second term implements the contrastive training between the synthesized pseudo-positive QA pairs, $(\mathbf{q},\mathbf{a}^*)$ and $(\mathbf{q}^*,\mathbf{a}^-)$, and the original negative QA pairs, $(\mathbf{q},\mathbf{a}^-)$. 

\section{Experiment}
\subsection{Experimental Setup}
\subsubsection{\textbf{Datasets and Evaluation Metrics}} 
We evaluate our method on three widely-used benchmark datasets for answer selection: TREC-QA, WikiQA, and ANTIQUE. The statistics of these datasets are described in Table~\ref{dataset}.  
\begin{itemize}[leftmargin=*]
    \item TREC-QA~\cite{DBLP:conf/emnlp/WangSM07}, collected from TREC QA track 8-13 data, is a widely-adopted benchmark for factoid question answering. 
    \item WikiQA~\cite{DBLP:conf/emnlp/YangYM15} is an open-domain factoid answer selection benchmark, in which the answers are collected from the Wikipedia. 
\end{itemize}
Following previous studies~\cite{tanda}, we adopt the RAW version of training data and evaluate on the cleaned test data for both TREC-QA and WikiQA. The mean average precision (MAP) and mean reciprocal rank (MRR) are adopted as the evaluation metrics.
\begin{itemize}[leftmargin=*]
    \item ANTIQUE~\cite{antique} is an open-domain non-factoid QA dataset collected from community question answering services with a diverse set of categories. 
\end{itemize}
Although there are four-level relevance labels for the candidate answers, we follow previous studies~\cite{antique,sigir20-curriculum} to assume that the label 3 and label 4 are relevant, while 1 and 2 are non-relevant. MRR and P@1 are adopted for evaluation on the binary labels, while nDCG is adopted for evaluation on the four-level relevance labels. 

\begin{table}
\setlength{\abovecaptionskip}{2pt}   
\setlength{\belowcaptionskip}{2pt}
\centering
  \begin{adjustbox}{max width=\linewidth}
  \begin{tabular}{c|c|c|c}
    \toprule
	Dataset & \multirow{2}{*}{\#Question} & \multirow{2}{*}{\#QA Pairs} &\multirow{2}{*}{\%Positive} \\
    (train/dev/test) &&& \\
    \midrule
   TREC-QA & 1160/65/68 & 53417/1117/1442 & 12.0/18.4/17.2\\
    WikiQA& 2118/126/243 &20360/1130/2351 &12.0/12.4/12.5 \\
    ANTIQUE & 2426/-/200 & 27422/-/6589& 72.2/-/38.4\\
  \bottomrule
\end{tabular}
\end{adjustbox}
\caption{Summary statistics of datasets.}\label{dataset}
\vspace{-0.5cm}
\end{table}

\subsubsection{\textbf{Baseline Methods}}
Since TREC-QA and WikiQA have been extensively studied for answer selection, we are able to compare with more diverse baseline methods, including traditional methods and Transfer Learning based methods.
\begin{itemize}[leftmargin=*]
    \item We consider the methods without using external resources during the training phase as traditional methods.  aNMM~\cite{anmm}, RNN-POA~\cite{sigir17-poa}, and KABLSTM~\cite{sigir18-kablstm} develop different kinds of attention-based structures for enhancing the interaction between the question and answer. HyperQA~\cite{hyperqa} and SD~\cite{sigir20-as} study the similarity aggregation approaches in siamese neural network structure. Currently, models based on pre-trained contextualized representations, e.g., BERT~\cite{bert}, achieve state-of-the-art performance among the traditional methods.
    \item TL-based methods refer to those methods that transfers cross-domain knowledge from external large-scale resources to enhance the resource-poor tasks. Here we compare our results with two state-of-the-art TL-based methods, i.e.,  CA+LM+LC+TL~\cite{cikm19-ca} and TANDA~\cite{tanda}, which adopts QNLI (86K QA pairs) and ASNQ (20M QA pairs) as source dataset, respectively, for transfer learning. 
\end{itemize}

As for ANTIQUE, we compare to the benchmark performance on BERT and a newly developed curriculum learning strategy~\cite{sigir20-curriculum} for learning to handle difficult cases. 

\subsubsection{\textbf{Implementation Details}}
We adopt the Large version of BART~\cite{bart} as the conditional generation model for data augmentation. 
During the fine-tuning of the question generator and answer generator, we set the maximum sequence length for question and answer as 30 and 50, respectively. 
Both of the question generator and answer generator are fine-tuned for 5 epochs. 
All other hyper-parameters are set to default as the original BART.
Since the proposed method can be applied to any ranking model, we choose two widely-adopted ranking models, namely, BERT~\cite{bert} and aNMM~\cite{anmm}, as the base models for evaluation. 
The maximum sequence length of the concatenated QA pair is set to be 128. We train all the models for 5 epochs. 
All other hyper-parameters are also set to default as the original BERT. And we use MatchZoo~\cite{matchzoo} to implement the aNMM model.

\subsection{Overall Performance}

\begin{table}
\centering
\setlength{\abovecaptionskip}{2pt}   
\setlength{\belowcaptionskip}{2pt}

\begin{adjustbox}{max width=\linewidth}
\setlength{\tabcolsep}{1mm}{
\begin{tabular}{lcccc}
\toprule 
\multirow{2}{*}{Method}& \multicolumn{2}{c}{TREC-QA}& \multicolumn{2}{c} {WikiQA}  \\
\cmidrule(lr){2-3} \cmidrule(lr){4-5} 
& MAP & MRR & MAP &MRR \\
\midrule
aNMM&0.750& 0.811&0.610&0.628\\
RNN-POA&0.781&0.851 &0.721& 0.731\\
KABLSTM&0.804& 0.885& 0.732& 0.749\\
HyperQA&  0.784& 0.865&0.712& 0.727\\
SD &0.783&0.878&0.704&0.712\\
BERT$_\mathrm{base}$&0.857&0.937&0.813&0.828\\
BERT$_\mathrm{large}$&0.904&0.946&0.836&0.853\\
\midrule
CA+LM+LC+TL(QNLI) &0.875&0.940&0.834&0.848\\
BERT$_\mathrm{base}$ TANDA(QNLI)&0.863&0.906&0.832&0.852\\
BERT$_\mathrm{base}$ TANDA(ASNQ)&0.912&0.951&0.893&0.903\\
BERT$_\mathrm{large}$ TANDA(ASNQ)&0.912&0.967&0.904&0.912\\
\midrule
\midrule
\textbf{aNMM-BiG}&0.792$^\dagger$&0.879$^\dagger$&0.698$^\dagger$&0.710$^\dagger$\\
\textbf{BERT$_\mathrm{base}$-BiG}&\textbf{0.899}$^\dagger$&\textbf{0.961}$^\dagger$&\textbf{0.825}$^\dagger$&\textbf{0.836}$^\dagger$\\
\textbf{BERT$_\mathrm{large}$-BiG}&\textbf{0.913}$^\dagger$&\textbf{0.966}$^\dagger$&\textbf{0.855}$^\dagger$&\textbf{0.863}$^\dagger$\\
\bottomrule
\end{tabular}}
\end{adjustbox}
\caption{Experimental Results on TREC-QA and WikiQA. $^\dagger$ indicates significant improvement ($p<0.05$) over the corresponding base ranking models.}
\label{trecwiki}
\vspace{-0.5cm}
\end{table}

Table~\ref{trecwiki} summarizes the method comparisons on TREC-QA and WikiQA. 
Here we combine our BiG method with both an attention-based ranking model, namely aNMM, and the pre-trained BERT ranking model. Two groups of methods on answer selection, in terms of whether to use large-scale external resources, are adopted for comparisons. 

Among \textbf{traditional methods}, models using contextualized representations, i.e., BERT$_\mathrm{base}$ and BERT$_\mathrm{large}$, achieve distinguishable state-of-the-art performance, showing the power of pre-trained languge models. 
All the ranking models combined with the proposed data augmentation strategy, BiG, consistently and substantially improve the performance of the original ranking models. 
For instance, BERT$_\mathrm{base}$-BiG outperforms the original BERT$_\mathrm{base}$ by about 5\% and 2\% on TREC-QA and WikiQA datasets, respectively.  
The results indicate that the proposed data augmentation strategy effectively synthesize valuable contrastive data for learning to rank question answer pairs without the need of large-scale annotated data.

As for \textbf{transfer learning based methods}, the performances on target tasks step up to the next stage with large-scale external resources. 
Compared with CA+LM+LC+TL(QNLI) and TANDA(QNLI), which employs 86K external annotated QA pairs from QNLI for training,  BiG-enhanced models achieve a competitive performance by only manipulating the limited original data. 
Even if comparing to TANDA(ASNQ), which employs 20M external QA pairs from Natural Questions for training, BiG-enhanced models can still outperform it on the MRR metric of TREC-QA. 
However, since Natural Questions and WikiQA are both derived from Wikipedia, they are likely to share a lot of related in-domain knowledge, leading to a great performance boosting on WikiQA.

Table~\ref{antique} presents the experimental results on ANTIQUE.
In general, BiG significantly improves the performance of BERT$_\mathrm{base}$ by about 6\%. Especially for nDCG, we observe a large performance increase in the multi-level ranking metrics. This result shows that the proposed method is also effective on non-factoid QA.  

\begin{table}
\centering
\setlength{\abovecaptionskip}{2pt}   
\setlength{\belowcaptionskip}{2pt}
\begin{adjustbox}{max width=\linewidth}
\setlength{\tabcolsep}{1mm}{
\begin{tabular}{lcccccc}
\toprule 
Method & MRR & P@1 & nDCG@1 &nDCG@3 &nDCG@10 \\
\midrule
aNMM&0.625&0.485&0.529&0.513&0.490\\
BERT$_\mathrm{base}$&0.797&0.709&0.713&0.657&0.642\\
BERT$_\mathrm{large}$&0.802&0.690&0.708&0.669&0.693\\
BERT$_\mathrm{base}$ Pair.&0.700&0.585&-&-&-\\
~ - w/ $\mathcal{D}_{recip}$&0.734&0.645&-&-&-\\
\midrule
\midrule
\textbf{aNMM-BiG}&0.662$^\dagger$&0.565$^\dagger$&0.601$^\dagger$&0.586$^\dagger$&0.589$^\dagger$\\
\textbf{BERT$_\mathrm{base}$-BiG}&\textbf{0.847}$^\dagger$&\textbf{0.765}$^\dagger$&\textbf{0.750}$^\dagger$&\textbf{0.710}$^\dagger$&\textbf{0.720}$^\dagger$\\
\textbf{BERT$_\mathrm{large}$-BiG}&\textbf{0.859}$^\dagger$&\textbf{0.785}$^\dagger$&\textbf{0.772}$^\dagger$&\textbf{0.719}$^\dagger$&\textbf{0.738}$^\dagger$\\
\bottomrule
\end{tabular}}
\end{adjustbox}
\caption{Experimental Results on ANTIQUE.}
\label{antique}
\vspace{-0.6cm}
\end{table}

\subsection{\textbf{Ablation Study}}
According to the ablation studies in Table~\ref{ablation}, we observe that both the BiG data augmentation and the contrastive learning objective contribute to the final performance more or less.

As for TREC-QA and ANTIQUE, pairwise training achieves a better performance than pointwise training. Since there is a large number of questions without any positive answer in the WikiQA dataset, the lack of pairwise training samples causes that the performance of pairwise training is much worse than pointwise training. 
This issue is appropriately solved by our data augmentation strategy by constructing synthesized pseudo-positive samples. 

The QG-augmented data contributes much more to the performance on WikiQA than AG, while the AG-augmented data is slightly more beneficial to TREC-QA and ANTIQUE than QG.  
Since there are only 873 out of 2118 questions containing positive answers in the training data of WikiQA, the rest questions are all underutilized for pairwise training. The synthesized contrastive data by QG effectively evoke such valuable knowledge for learning.

Compared with pointwise training with only the original data, ``w/o contrastive learning", where the synthesized QA pairs are regarded as ground-truth positive samples for pointwise training, can achieve relatively better performance. However, there is still a great margin to reach the performance of BERT$_\mathrm{base}$-BiG. The results indicate that the synthesized QA pairs are actually more relevant than the original negative samples, but the proposed contrastive learning objective can better utilize the synthesized data than treating them as positive samples for pointwise training. 

\begin{table}
\centering
\setlength{\abovecaptionskip}{2pt}   
\setlength{\belowcaptionskip}{2pt}
\begin{adjustbox}{max width=\linewidth}
\setlength{\tabcolsep}{1mm}{
\begin{tabular}{lcccccc}
\toprule 
\multirow{2}{*}{Method}& \multicolumn{2}{c}{TREC-QA}& \multicolumn{2}{c} {WikiQA} & \multicolumn{2}{c} {ANTIQUE} \\
\cmidrule(lr){2-3} \cmidrule(lr){4-5} \cmidrule(lr){6-7} 
& MAP & MRR & MAP &MRR & MRR & P@1\\
\midrule
BERT$_\mathrm{base}$ Point.&0.857&0.937&0.813&0.828&0.797&0.709\\
BERT$_\mathrm{base}$ Pair.&0.872&0.925&0.734&0.747&0.806&0.715\\
\textbf{BERT$_\mathrm{base}$-BiG}&\textbf{0.899}&\textbf{
0.961}&\textbf{0.825}&\textbf{0.836}&\textbf{0.847}&\textbf{0.765}\\
\midrule 
- w/o QG&0.891&0.958&0.758&0.772&0.844&0.755\\
- w/o AG&0.882&0.949&0.824&0.833&0.834&0.740\\
- w/o Contrastive&0.878&0.942&0.818&0.831&0.802&0.690\\
\bottomrule
\end{tabular}}
\end{adjustbox}
  \caption{Ablation Study.}
\label{ablation}
\vspace{-0.6cm}
\end{table}

\section{Conclusions}
In this paper, we study the data augmentation for improving the performance of ranking question answer pairs when encountering data scarceness and label imbalance issues. 
We propose an easy-to-apply and effective data augmentation strategy, namely Bilateral Generation, with a contrastive learning objective to maximize the utility of limited labeled data. 
Experimental results show that the proposed method significantly improves the performance of ranking models on three factoid and non-factoid QA benchmark datasets. 

The proposed data augmentation strategy can be easily extended to few-shot learning scenario to further handle more severe resource-poor applications, such as domain-specific question answering, e-Commerce question answering, etc.

\bibliography{anthology,custom}
\bibliographystyle{acl_natbib}

\end{document}